\documentclass{article}

\usepackage{hologo}

\usepackage{fontawesome}

\usepackage[final]{neurips_2025}

\usepackage{natbib}
\setcitestyle{numbers,square}
\usepackage[utf8]{inputenc} 
\usepackage[T1]{fontenc}    
\usepackage{hyperref}       
\usepackage{url}            
\usepackage{booktabs}       
\usepackage{amsfonts}       
\usepackage{nicefrac}       
\usepackage{microtype}      
\usepackage{xcolor}         
\usepackage{multirow}
\usepackage{graphicx}
\usepackage[flushleft]{threeparttable}
\usepackage{adjustbox}
\usepackage{subcaption}
\usepackage{longtable}
\usepackage{booktabs} 
\usepackage{colortbl} 

\usepackage{hyperref}

\usepackage{caption}  
\usepackage{geometry}  
\usepackage{array}  

\title{WanJuanSiLu: A High-Quality Open-Source Webtext Dataset for Low-Resource Languages}

\author{%
Jia Yu \quad Fei Yuan \quad Rui Min \quad Jing Yu \quad Pei Chu \quad Jiayang Li\\
\textbf{Wei Li} \quad \textbf{Ruijie Zhang} \quad \textbf{Zhenxiang Li} \quad \textbf{Zhifei Ren} \quad \textbf{Dong Zheng} \\
\textbf{Wenjian Zhang} \quad \textbf{Yan Teng} \quad \textbf{Lingyu Meng} \quad \textbf{ZhenJiang Jin} \quad \textbf{Jiantao Qiu} \\
\textbf{ShaSha Wang} \quad \textbf{Zhongying Tu} \quad \textbf{Dahua Lin} \quad \textbf{Yu Wang} \\
\textbf{Yu Qiao}\footnotemark[2] \quad \textbf{Yanfeng Wang}\footnotemark[2] \quad \textbf{Conghui He}\footnotemark[2] \\
Shanghai Artifcial Intelligence Laboratory\\
Shanghai, 200232, China\\
\texttt{ \{qiaoyu, wangyanfeng, heconghui\}@pjlab.org.cn } \\
}

\begin{document}

\maketitle

\begin{abstract}
This paper introduces the open-source dataset WanJuanSiLu, designed to provide high-quality training corpora for low-resource languages, thereby advancing the research and development of multilingual models. To achieve this, we have developed a systematic data processing framework tailored for low-resource languages. This framework encompasses key stages such as data extraction, corpus cleaning, content deduplication, security filtering, quality evaluation, and theme classification. Through the implementation of this framework, we have significantly improved both the quality and security of the dataset, while maintaining its linguistic diversity. As of now, data for all five languages have been fully open-sourced. The dataset can be accessed at \href{https://opendatalab.com/applyMultilingualCorpus}{{https://opendatalab.com/applyMultilingualCorpus}}, and GitHub repository is available at \href{https://github.com/opendatalab/WanJuan3.0}{{https://github.com/opendatalab/WanJuan3.0}}
\end{abstract}

\section{Introduction}

Large Language Models (LLMs)\cite{8,2,achiam2023gpt}, with their powerful natural language processing capabilities, are reshaping the service paradigms of artificial intelligence applications. However, current mainstream LLMs primarily focus on a few high-resource languages, leaving the majority of users among the world's 7,000+ languages unable to fully benefit from AI technologies. This uneven technological coverage not only exacerbates disparities in information access but also hinders computational linguistics' comprehensive understanding of human language systems.

To address this issue, promoting the development of LLMs for low-resource languages has become an urgent priority. The challenges in developing LLMs for low-resource languages are particularly pronounced: first, the limited corpus size directly restricts the generalization ability of the models; second, the scattered nature of data sources and regional differences significantly increase the complexity of building standardized corpora; and finally, the lack of technical support and linguistic resources poses additional hurdles for multilingual corpus processing.

To tackle these challenges, we have developed a systematic multilingual data processing framework, with enhancements in data labeling and data security. For data labeling, we proposed a classification system tailored to the distributional characteristics of multilingual corpora and domain knowledge structures. This system adopts a two-tier classification architecture: the primary level covers eight core domains, including encyclopedic knowledge, local life, and cultural knowledge, while the secondary level further refines these into 34 subcategories.

\newpage

In summary, the main contributions of this work include:
\begin{enumerate}
    \item We have constructed and open-sourced a large-scale multilingual text corpus, \textbf{WanjuanSiLu}, which includes five languages, over 350 million documents, 1.2TB of data, and more than 300 billion tokens. The corpus is categorized into 7 major categories and 34 subcategories.
    \item We proposed a refined data processing workflow in Chapter\ref{sec:method}, which includes data cleaning, quality screening, security filtering, and thematic classification. This workflow effectively integrates multilingual features with industry-specific knowledge. Not only does it ensure the diversity and quality of the data, but it also establishes a comprehensive security mechanism, providing reliable data support for multilingual model training.

\end{enumerate}

\section{Dataset Statistics}
WanjuanSiLu as a comprehensive multilingual text corpus, has collected public network information, literature, patents, and other materials from multiple countries and regions. The total data size exceeds 1.2TB, with a token count of over 300 billion. The first phase of the open-source corpus mainly consists of five subsets: Thai, Russian, Arabic, Korean, and Vietnamese, with each subset having data scale of over 150GB. The data volume statistics for the open-source corpus are shown in the Table \ref{tab:dataset_statis} .

\begin{table}[h!]
\centering
\renewcommand{\arraystretch}{1.5} 
\setlength{\tabcolsep}{3pt} 
\caption{{Dataset Statistics of the Open-Source Corpus}}
\vspace{0.1cm}
\begin{tabular}{cccc}
\hline
\textbf{lang} & \textbf{row} & \textbf{bytes (GB)} & \textbf{Token (B)} \\
\hline
ar & 123323645 & 224.8 & 57.2 \\
ko & 68894936  & 278.2 & 87   \\
ru & 60206389  & 302   & 62.6 \\
th & 23391887  & 159.9 & 31.8 \\
vi & 80970018  & 280.2 & 74.1 \\
\hline
\textbf{Total} & 356786875 & 1245.2 & 312.7 \\
\hline
\end{tabular}

\label{tab:dataset_statis}
\end{table}


\begin{figure}[htbp] 
    \centering
    \includegraphics[width=10cm]{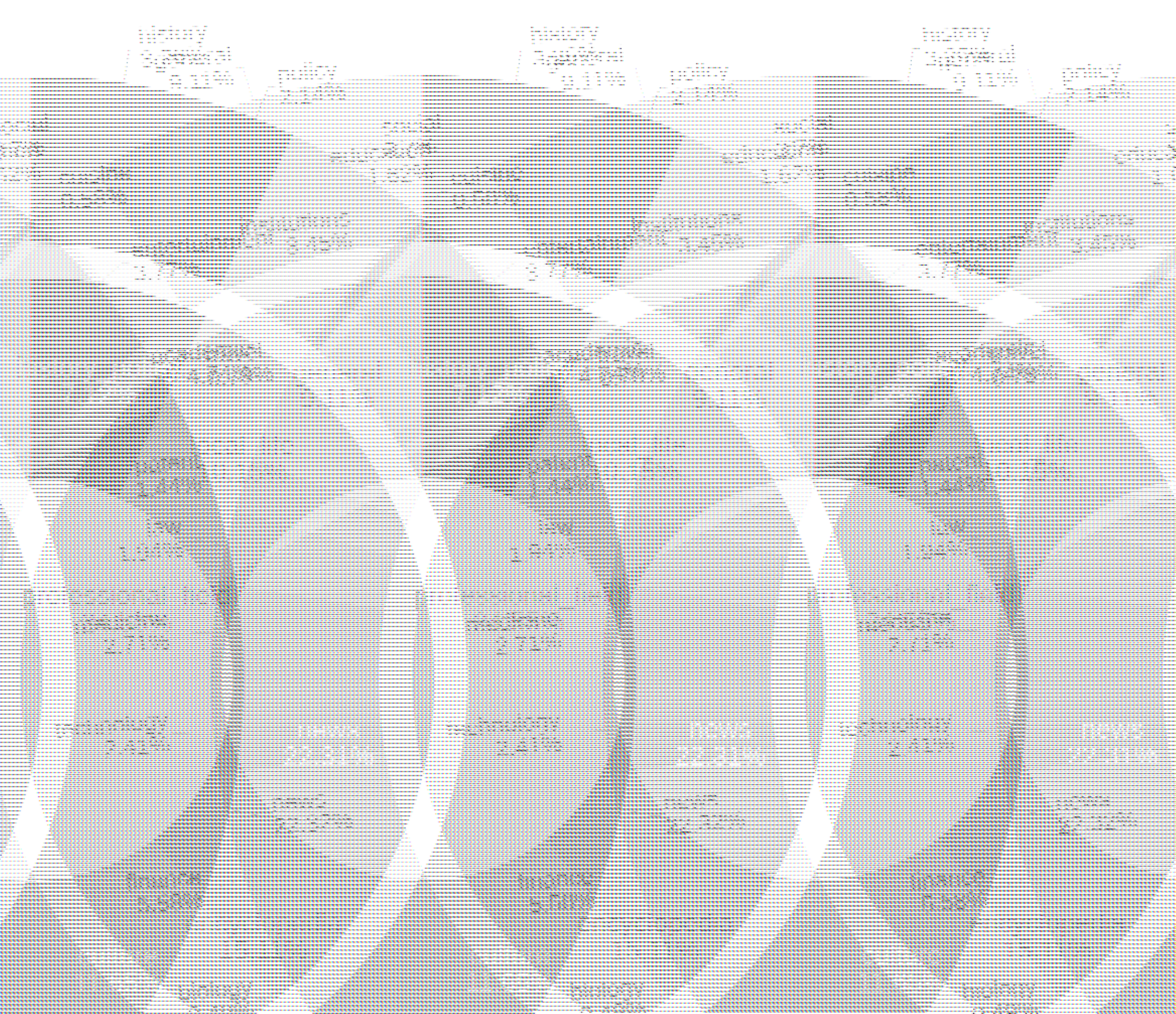} 
    \caption{{Distribution of Themes Across Level 1 and Level 2 Labels in the Corpus}} 
    \label{fig:theme} 
\end{figure}

\newpage

Based on the Classification System, we divided the corpus into 7 major categories and 34 subcategories, covering a wide range of content with characteristics of the language's geographic location, such as history, politics, culture, real estate, shopping, weather, dining, encyclopedic knowledge, and specialized knowledge. The distribution of the first and second-level labels in the corpus is shown in Figure \ref{fig:theme}.


To comprehensively evaluate processing methods and features of existing datasets, we conducted a multi-dimensional comparative study on representative multilingual public datasets such as MADLAD-400 \cite{4}, CC100 \cite{2}, and mC4 \cite{1}. The comparative analysis mainly focused on four core dimensions: data cleaning methods, quality evaluation strategies, security processing mechanisms, and topic classification systems. The specific comparison results are shown in Table \ref{tab:dataset_comparison}.

\begin{table}[h!]
\centering
\scriptsize 
\renewcommand{\arraystretch}{1.5} 
\setlength{\tabcolsep}{3pt} 
\caption{{Comparison of Dataset Filtering Strategies}}
\begin{tabular}{|c|c|c|c|c|c|}
    \hline
    \textbf{Dataset} &  \textbf{Rule-based clean} & \textbf{Quality Filtering-PPL}  & \textbf{Quality Filtering-Cls $^*$} & \textbf{Safety Filtering-Cls $^*$} & \textbf{Theme Classification}\\ \hline
    MADLAD-400 & {\color{red}Yes} & No & No & No & No\\ \hline
    CC100 &{\color{red}Yes}  & {\color{red}Yes} & No & No & No \\ \hline
    mC4 & {\color{red}Yes} & No & No & No & No \\ \hline
    WanJuanSiLu & {\color{red}Yes} & {\color{red}Yes} & {\color{red}Yes} &{\color{red}Yes} & {\color{red}Yes} \\ \hline
\end{tabular}
 \begin{tablenotes}   
    \footnotesize              
    \item[1] $^*$ : Quality Filtering-Cls and Safety Filtering-Cls mean using model-based classifier to filter the data         
  \end{tablenotes}         
\vspace{0.1cm}

\label{tab:dataset_comparison}
\end{table}

\section{Method}
\label{sec:method}
When processing large scale low-resource language data, existing general preprocessing methods often face multiple challenges: significant language feature differences, inconsistent standardization, and difficulties in quality evaluation. To address this, we have designed a systematic data processing workflow tailored to the characteristics of low-resource languages. By combining linguistic expert knowledge and automated processing techniques, we have significantly improved the overall quality of low-resource language data through multi-stage refined processing.

\begin{figure}[htbp] 
    \centering
    \includegraphics[width=12.5cm]{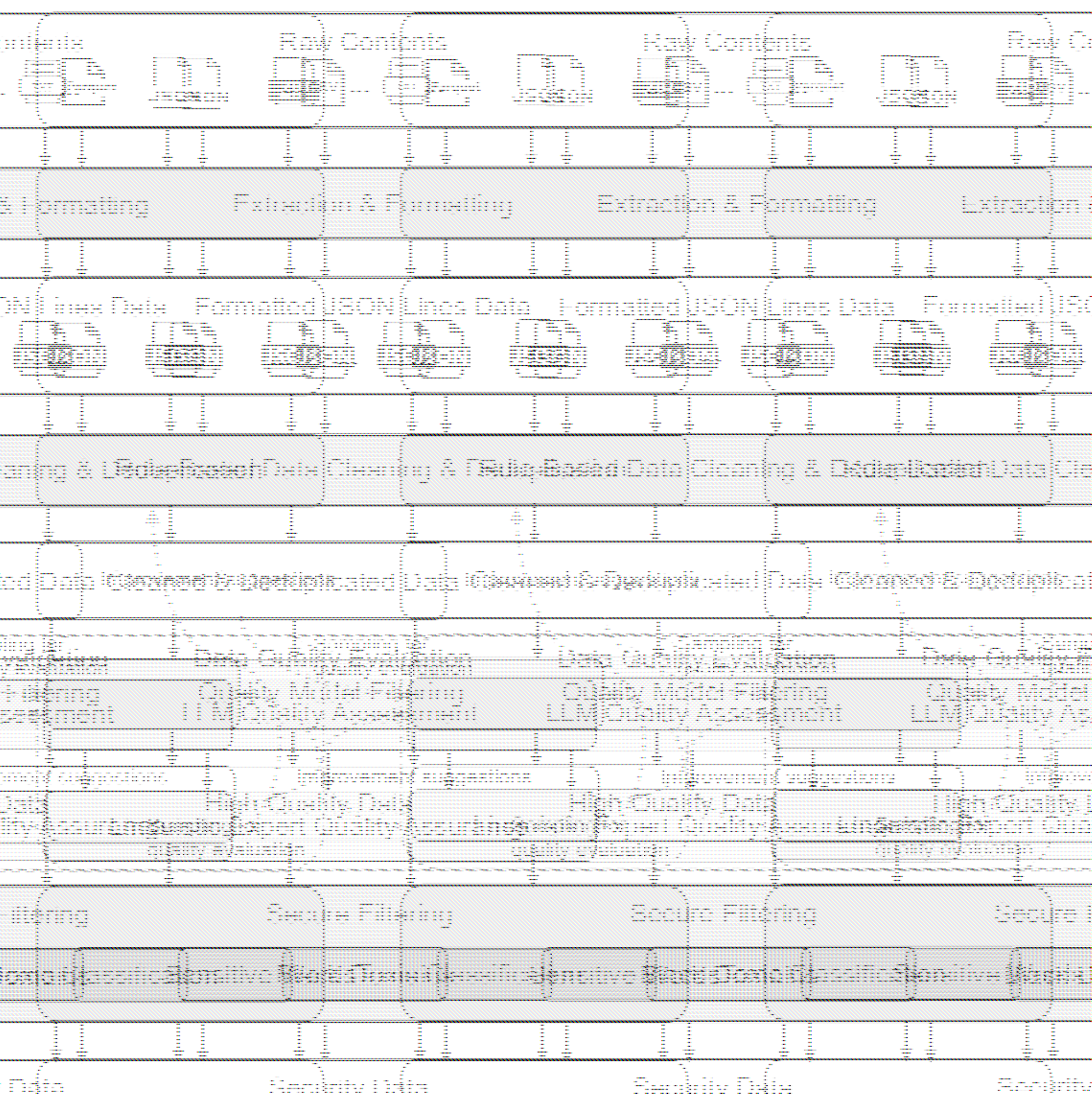} 
    \caption{Data Processing Pipeline} 
    \label{fig:data_process_pip} 
\end{figure}

As shown in the Figure \ref{fig:data_process_pip}, our data processing workflow includes the following core steps: data extraction \& formatting, rule-based cleaning \& deduplication, quality evaluation, security filtering, and topic classification. Focusing on the characteristics of low-resource language data, we have strengthened the following key aspects:
\begin{itemize}
    \item Customizing specialized data cleaning rules for different languages based on linguistic features.
    \item Combining linguistic expert quality inspection with automated evaluation to establish a differentiated quality control system.
    \item Building a multi-level security processing mechanism, including language-specific sensitive word and security model filtering.    
\end{itemize}

\subsection{Data Clean}
The data cleaning process is applied to web-based HTML data, including sources such as Common Crawl and other general datasets.

During the content extraction and standardization phase, we employ a self-developed text extraction tool to extract the core content from raw web pages and standardize it into the JSON Lines format. Subsequently, a language detection model based on FastText \cite{fasttext, fasttext2} is used to identify the language of each document, retaining only data in the target languages.

In the rule-based cleaning and deduplication phase, we applied multi-dimensional heuristic rules to systematically clean the corpus. This process primarily addresses several typical noise issues, including text fragments with insufficient length, residual HTML tags, and non-standard text terminators. Considering the specific characteristics of multilingual corpora, we also specifically handled detail issues that might affect formatting, such as invisible characters, abnormal spaces around punctuation marks, and special characters in certain languages.

To ensure the comprehensiveness of data quality, we designed a two-layer evaluation mechanism. After completing the initial data cleaning, we adopted a batch random sampling strategy, selecting 10,000 samples from each batch to be sent to a large language model for quality evaluation and scoring. Then, based on the scoring distribution of the sample data, we extracted an appropriate amount of data from both high and low-scoring samples at a reasonable ratio and sent them to language domain experts for in-depth quality inspection. For the data output by the quality model, we applied the same quality evaluation process.

Finally, we employ a MinHash-based fuzzy deduplication method \cite{deduplicatingtrainingdatamakes,repeat,minhash} to reduce redundancy in the corpus.

\subsection{Quality Evaluation}
In the field of data processing, quality Evaluation is a key step, and this is particularly emphasized in the processing of low-resource language data. To establish a scientific evaluation system, we have developed a multi-level quality classification framework that systematically categorizes data issues into three main dimensions: quality, duplication, and security which are shown in Table \ref{tab:specific_problem_definition} . Due to its inherent complexity, quality issues are further subdivided into five evaluation sub-dimensions: relevance, completeness, understandability, effectiveness, and fluency.

\begin{table}[h!]
\centering

\small 
\renewcommand{\arraystretch}{1.5} 
\setlength{\tabcolsep}{5pt} 
\caption{{Classification and Definition of Specific Problems}}
\resizebox{1.01\linewidth}{2.5cm}{
\begin{tabular}{cccp{7.7cm}}
\hline
\textbf{No.} & \textbf{Primary Category} & \textbf{Secondary Category} & \multicolumn{1}{c}{\textbf{Definition}} \\ 
\hline
1 & \multirow{5}{*}{Quality Issues} & Completeness & Whether the data content is semantically complete \\ 
\arrayrulecolor[rgb]{0.8, 0.8, 0.8} 
\cline{1-1} \cline{3-3} \cline{4-4}
2 &  & Validity & Whether the data content contains valid semantic information \\ 
\cline{1-1} \cline{3-3} \cline{4-4}
3 &  & Comprehensibility & Whether the data content is ambiguous or incomprehensible due to errors such as formatting issues \\ 
\cline{1-1} \cline{3-3} \cline{4-4}
4 & & Fluency & Whether the data content is semantically fluent \\ 
\cline{1-1} \cline{3-3} \cline{4-4}
5 &  & Relevance & Whether the data contains content relevant to the context or main topic \\ 
\hline
6 & Duplication Issues & Similarity & Whether the data is duplicated \\ 
\hline
7 & Security Issues & Security & Whether the data involves content safety concerns \\
\arrayrulecolor{black} 
\hline
\end{tabular}
}

\label{tab:specific_problem_definition}
\end{table}

\begin{table}[h!]
\centering
\small 
\renewcommand{\arraystretch}{1.2} 
\setlength{\tabcolsep}{2pt} 
\caption{Manual Quality Inspection Issues $^*$}
\begin{tabular}{cccc}
\hline
\textbf{Category} & \textbf{Issue Description} & \textbf{Count} & \textbf{Proportion} \\ 
\hline
High Quality & High Quality & 11433 & 44.53\% \\ 

Quality - Relevance & Contains irrelevant content & 8598 & 33.49\% \\ 

Quality - Completeness & Content is incomplete & 3469 & 13.51\% \\ 

Duplication - Similarity & Content has duplication issues & 1335 & 5.20\% \\ 

Safety - Security & Content involves security issues & 1152 & 4.49\% \\ 

Quality - Comprehensibility & Content is incomprehensible & 1147 & 4.47\% \\ 

Quality - Validity & Contains invalid content & 1092 & 4.25\% \\ 

Quality - Fluency & Content is not fluent & 599 & 2.33\% \\ 
\hline
\end{tabular}

\vspace{0.01cm}
\begin{tablenotes}   
    \footnotesize   
    \centering
    \item[1] $^*$ : A single data entry may involve multiple issues, which will be counted repeatedly in total number of issues        
\end{tablenotes}


\label{tab:manual_problem}
\end{table}

We conducted a sampling analysis of the standardized corpus data, with a sample size over 20,000, and discovered several issues in the corpus: irrelevant content, incomplete data, low understandability, formatting anomalies, and lack of fluency. These issues may adversely affect the pre-training performance of large-scale language models. 

Therefore, we conducted a statistical analysis of the quality issues present in the sample data, with specific results shown in the Table \ref{tab:manual_problem}. From the statistical results, it can be seen that high-quality data accounts for only 44.53\%, with relevance issues (33.49\%) being the most prominent among the various quality problems, followed by completeness issues (13.51\%). While other types of issues have relatively smaller proportions, they should not be overlooked.

\begin{figure}[htbp] 
    \centering
    \includegraphics[width=14cm]{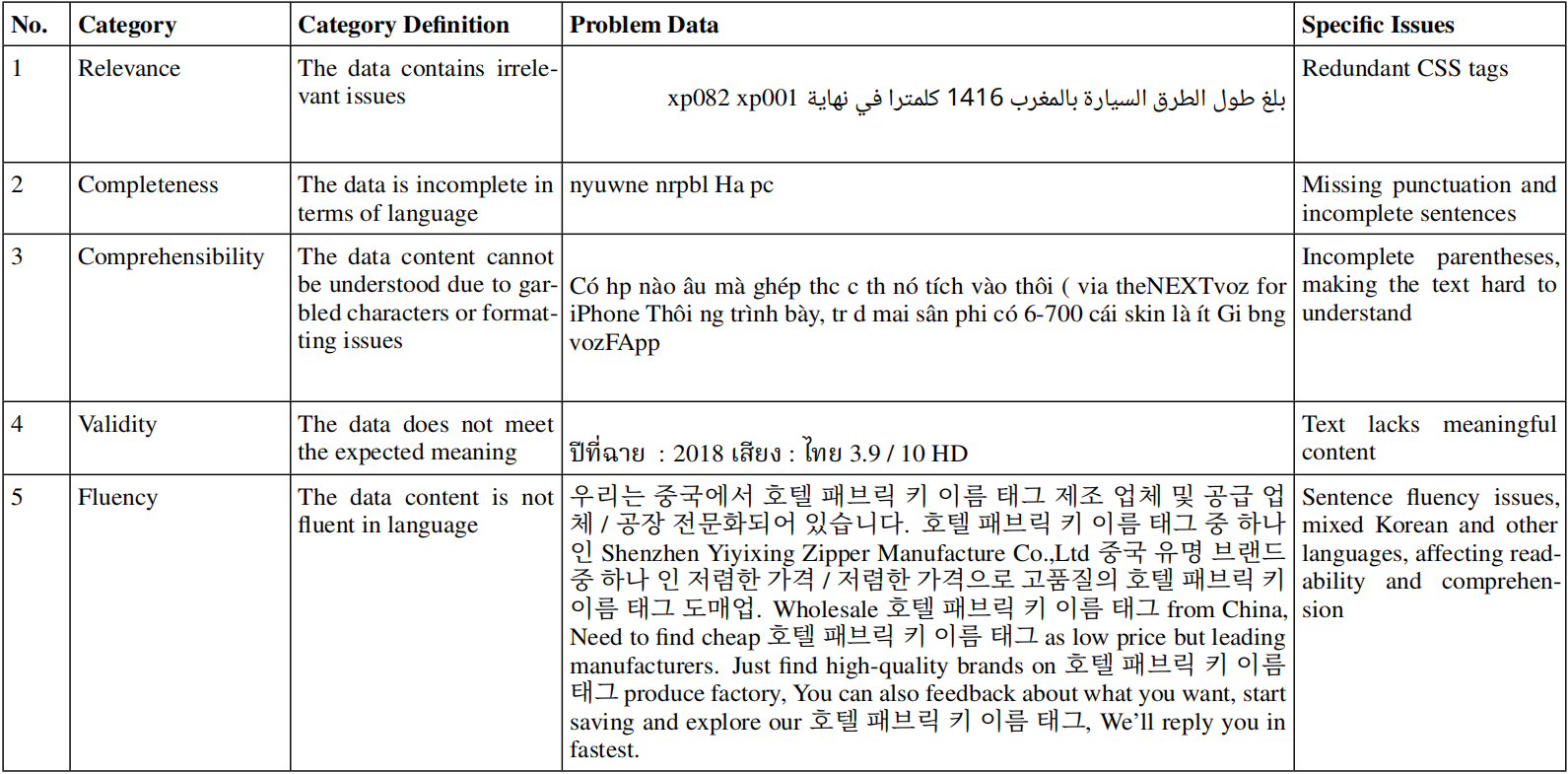} 
    \caption{{Examples of Quality Issues}} 
    \label{fig:quality_example} 
\end{figure}

To more intuitively understand the specific manifestations of quality issues, the Figure \ref{fig:quality_example} lists several typical examples that demonstrate the characteristics of different types of problems.

To ensure data quality and enhance data diversity, we designed a two-stage quality evaluation scheme. In the first stage, we perform preliminary screening based on perplexity (PPL). Specifically, we calculate the PPL value between the target text and the Wikipedia (Wiki) reference corpus to assess text quality, with a lower PPL value indicating higher text quality. We use the open-source 5-gram language model as the evaluation tool, which strikes a good balance between expressive capability and computational efficiency. For cross-lingual evaluation, we use pre-trained 5-gram models in five languages \cite{kenlm}.

\begin{figure}[htbp] 
    \centering
    \includegraphics[width=11cm]{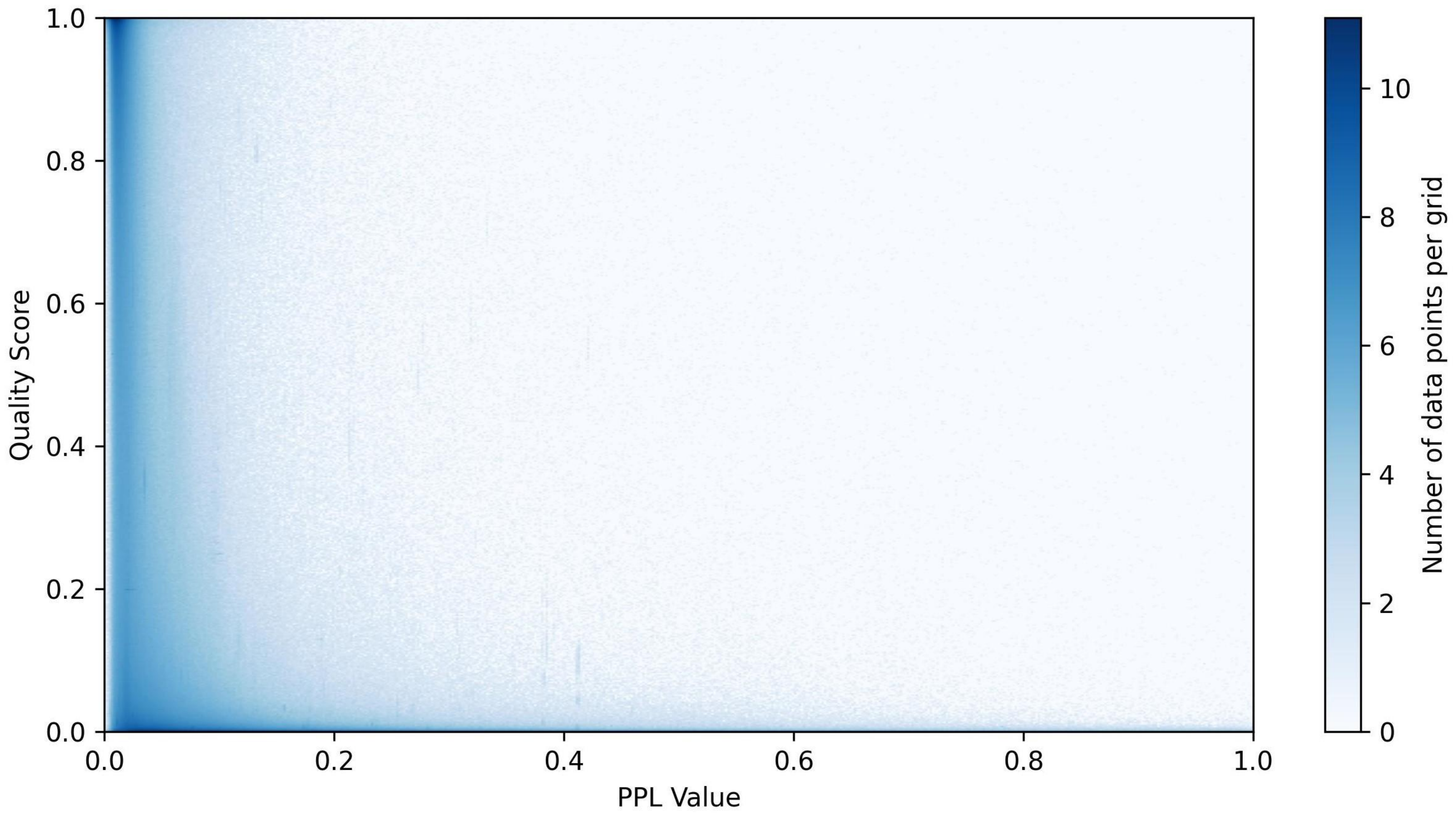} 
    \caption{{Diagram of Perplexity Values and Quality Scores}} 
    \label{fig:ppl} 
\end{figure}

Taking the Korean data as an example in Figure \ref{fig:ppl}, by analyzing the relationship between the PPL value and manual quality labels, we found that when the PPL value exceeds 0.2, the data mainly concentrates in the area with quality scores below 0.2. In contrast, high-quality data (with quality scores above 0.5) is primarily distributed in the PPL range from 0 to 0.2. Based on this finding, we set a screening threshold of a PPL value not exceeding 0.2, which effectively reduces the data size while retaining high-quality data. In practice, this method reduced the data volume by 37\% for Korean data, with reductions of 30\% to 50\% in other languages.

In the second stage, we developed a quality scoring model based on multilingual-BERT \cite{mbert} to conduct a refined evaluation of the data that passed the initial filtering, considering dimensions such as information content, data noise, and advertisement recognition. The model, through contrastive training on high-quality and low-quality data, is able to accurately identify subtle differences in corpus quality. Given the computational overhead of BERT models, we applied this model only to the candidate data that passed the initial evaluation. This strategy ensures high filtering accuracy while significantly improving processing efficiency. Through this two-stage screening scheme, we have ensured both the high quality of the data and the efficient use of computational resources.

\subsection{Safety Filtering}
In response to data security issues, we conducted a systematic analysis and processing. The security issues in multilingual corpora mainly manifest in uneven data distribution, significant differences in content severity, and widespread coverage of unsafe content. According to the statistics in Table \ref{tab:manual_problem}, unsafe content accounts for approximately 4.5\% of the overall corpus. These security risks could lead to the model learning inappropriate language patterns and knowledge representations, which could affect its practical application.

In terms of filtering strategy, we have developed a multi-level data security filtering system. By establishing and maintaining a domain blacklist, we achieve fast blocking of data from harmful websites. In sensitive word filtering, we developed an optimized multilingual filtering mechanism. Considering the specific characteristics of morphology and word formation rules in lesser-used languages, we used a combination of quantitative and qualitative methods to optimize the sensitive word list. Specifically, by calculating the inter-document frequency (IDF) and intra-document density of sensitive words, we identified high-priority evaluation targets, which, combined with the context evaluation by linguistic experts and content security specialists, effectively reduced the filtering misjudgment rate.

For detecting harmful content, we developed a multilingual multi-label detection model based on xlm-roberta-base \cite{xlmroberta}, which evaluates content in four dimensions: pornography, gambling, drugs, and violence. To improve the model's performance on low-resource language data, we took a series of measures: expanding training data using open-source samples, supplementing multilingual samples through cross-lingual translation, performing stratified sampling based on sensitive word density, and ensuring data quality through a combination of manual evaluation by language experts and machine annotations.

\begin{table}[h!]
\centering
\small 
\renewcommand{\arraystretch}{1.5} 
\setlength{\tabcolsep}{2pt} 
\caption{Performance of the Safety Model}
\vspace{0.1cm}
\begin{tabular}{cccc}
\hline
\textbf{Non-toxic Recall} & \textbf{Non-toxic Precision} & \textbf{Toxic Recall} & \textbf{Toxic Precision} \\ 
\hline
0.939 & 0.968 & 0.72 & 0.589 \\ 
\hline

\end{tabular}
\label{tab:performance_of_the_safety_model}
\end{table}

Tests showed in Table \ref{tab:performance_of_the_safety_model} that the model achieved recall and precision rates of 0.939 and 0.968 for benign samples, respectively, and recall and precision rates of 0.72 and 0.589 for harmful samples, demonstrating good detection performance. This security filtering system not only effectively identifies and filters harmful content but also maintains a low misjudgment rate, providing strong assurance for the safety of the model's training data.

\subsection{Theme Classification}
Given the general lack of thematic labeling in existing multilingual open datasets, we systematically classifies multilingual data into 7 primary themes and 34 secondary themes Table \ref{tab:labels} by analyzing the distribution characteristics of lesser-used language corpora and integrating domain knowledge structures. This structured thematic classification provides the model with multi-domain, organized corpus data. It provides the model with structured corpus data from multiple domains. To enhance accuracy and efficiency of data classification, we developed a multilingual topic classification model based on fastText \cite{fasttext_paper}. This model applies the classification framework to assign thematic labels to the processed corpus data.

\begin{table}[ht]
\centering
\small
\setlength{\arrayrulewidth}{0.01mm}
\arrayrulecolor[HTML]{404040} 
\renewcommand{\arraystretch}{1.3} 
\caption{The Classification System}
\begin{tabular}{p{2.5cm} p{10cm}}
\toprule
\multicolumn{1}{c}{\textbf{Level1}} & \multicolumn{1}{c}{\textbf{Level2}} \\ 
\midrule
\multirow{1}{*}{History \& Policy} & History; Heritage; Geography; Military; Political Policy  \\
 \hline
\multirow{1}{*}{News} & News  \\ \hline
\multirow{1}{*}{Culture} & Sports; Literature; Religion; Art; Customs; General Knowledge \\ \hline
\multirow{1}{*}{Encyclopedia} & Encyclopedia   \\  \hline
\multirow{2}{*}{Professional Field} & Finance; Technology; Medicine; Law; Patent; Academic; Institutions; Code; Education; Biology \\ \hline
\multirow{2}{*}{Local Life} & Social; Entertainment; Work; Office Tools; Life Tools; Real Estate; Cuisine; Weather; Shopping; Travel \\  \hline
General & General  \\ 
\bottomrule
\vspace{0.1cm}
\end{tabular}

\label{tab:labels}
\end{table}
In the system implementation process, we employed the Prompt technique of large language models to generate high-quality classification labels. This method enhances the system's ability to recognize specialized topics and improve classification accuracy. The system uses the FastText model as the core classification engine, enabling efficient processing of multilingual document inputs and delivering accurate thematic classification results based on the predefined classification system.

\section{Result}
To more comprehensively explain the production process and data quality of WanJuanSiLu, this chapter is divided into three parts for explanation: document retention analysis, comprehensive data scoring, and quality testing verification.

\begin{figure}[htbp] 
    \centering
    \includegraphics[width=15cm]{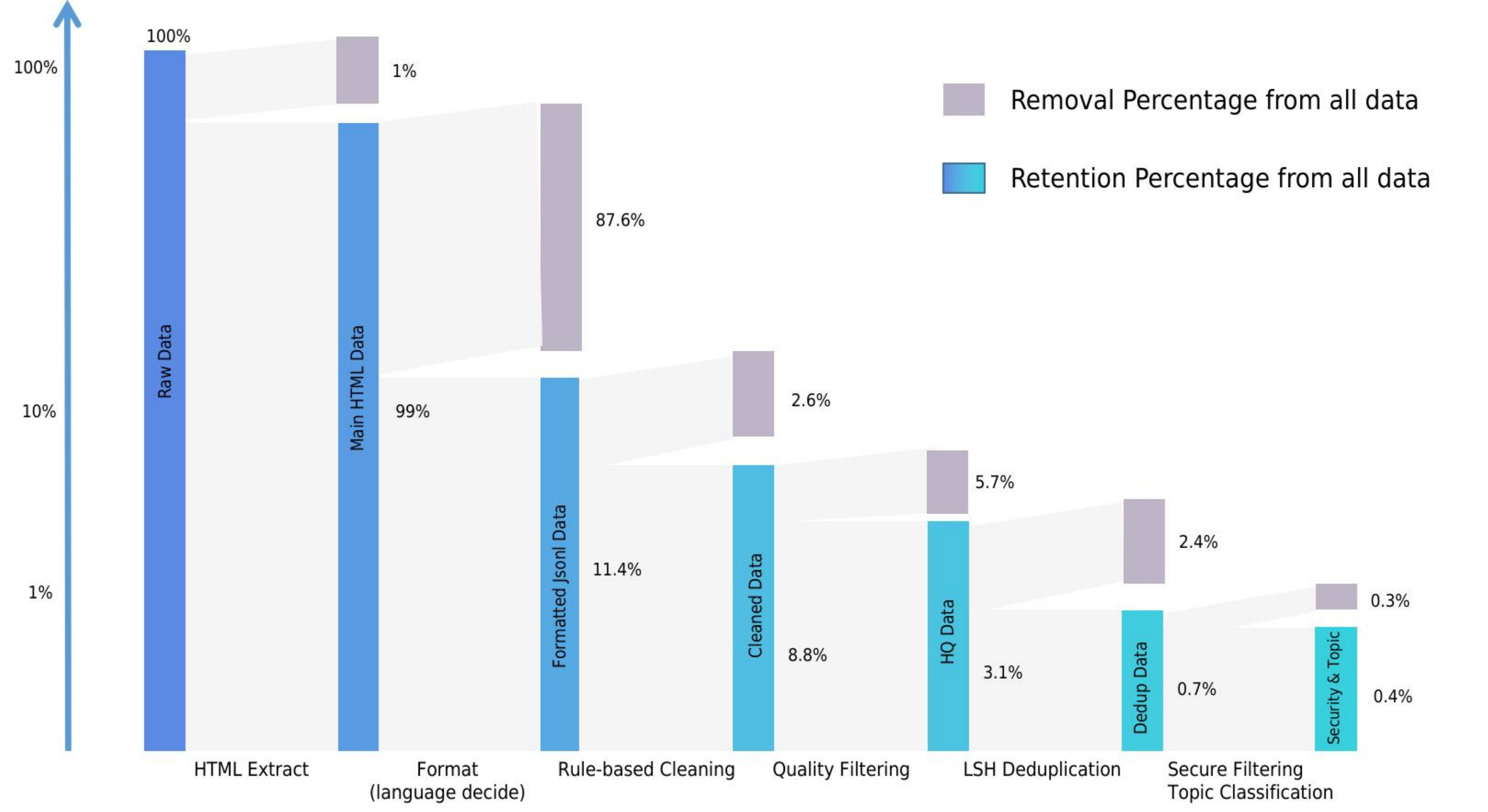} 
    \caption{Retention and Removal Rate for Different Stages} 
    \label{fig:pipline} 
\end{figure}

First, we calculated the data retention and deletion rates at each stage of the data processing pipeline, presenting these results in Figure \ref{fig:pipline}. Next, we utilized the open-source quality evaluation tool Dingo to perform a comprehensive assessment of the open-source dataset and compared its scores with those of other publicly available datasets. Finally, we detailed the results of model validation tests conducted using the open-source corpus.

\subsection{Analysis of Data Retention}

The changes in data volume at each stage of the data processing pipeline visually reflect the strictness of data quality control. By analyzing the dynamic changes in data retention rates, we can systematically observe the refined filtering process from the raw corpus to the high-quality dataset.

In the HTML extraction phase, to preserve the integrity of the original data, the data volume remained stable with a retention rate of 99\%. After entering the database, formatting, and target language identification filtering, the data retention rate significantly dropped to 11.4\%. This sharp decline reflects the small proportion of the target language in the overall corpus. The rule-based cleaning phase further filtered out 2.6\% of data that did not meet quality standards. Quality filtering reduced the data volume to 3.1\%, and LSH deduplication processing reduced the data volume to 0.7\%, effectively eliminating redundancy. Finally, after safety filtering and thematic classification, the data volume dropped to 0.4\%, ensuring both the security and thematic relevance of the dataset, thus constructing a high-quality target dataset.

\subsection{A Comprehensive Evaluation of Data Quality}
To comprehensively evaluate the effectiveness of the data cleaning process, we developed a systematic quality assessment framework. As part of this framework, a team of experts specializing in less commonly spoken languages conducted a manual inspection of the cleaned data, reviewing over 20,000 sample entries. Through detailed analysis, the team identified numerous quality issue indicators, which were subsequently categorized into seven core quality dimensions: completeness, validity, understandability, similarity, fluency, relevance, and security.

To achieve multi-dimensional data quality evaluation, we used the open-source evaluation tool dingo(\href{https://github.com/DataEval/dingo}{{https://github.com/DataEval/dingo}}), based on large language models, to perform comprehensive scoring on the corpus data. To verify the quality of the cleaned data, we selected three influential multilingual public datasets for comparative analysis. We evenly sampled data from each dataset, selecting 10,000 samples for each of five languages, and used the Qwen2.5-7B-Instruct model to provide a comprehensive score based on the seven dimensions. The scoring rules are as follows:
\begin{itemize}
  \item Each quality dimension is evaluated binary: when determined to be ``good," it receives 1 point, and when determined to be ``bad," it receives 0 points.
  \item Only when a data entry is determined to be ``good" in all seven dimensions, it is considered high-quality data and receives a comprehensive score of 1 point; otherwise, it receives 0 points.
\end{itemize}

Based on experimental data results in Table \ref{tab:dingo_score}, the WanJuanSiLu corpus leads in the comprehensive scores across all languages. The corpus achieved an average score of 95.74  in the evaluation of Arabic, Russian, Korean, Vietnamese, and Thai, demonstrating a high level of data quality. Notably, WanJuanSiLu performs consistently across all languages, with minimal score fluctuations. This characteristic highlights the corpus's advantage in cross-lingual data quality and security control.

\begin{table}[ht]
\centering
\renewcommand{\arraystretch}{1.2} 
\caption{{Comparison of the Evaluation Results on Different Datasets}}
\begin{tabular}{ccccc}
\hline
\textbf{language} & \textbf{WanJuanSiLu} & \textbf{MADLAD-400} & \textbf{CC100} & \textbf{mC4} \\
\hline
ar & 95.82 & 82.73 & 79.85 & 71.26 \\
ru & 96.78 & 73.40 & 84.24 & 62.02 \\
ko & 93.66 & 74.09 & 64.57 & 54.09 \\
vi & 97.33 & 82.12 & 72.15 & 64.56 \\
th & 95.11 & 53.06 & 63.55 & 53.89 \\
\textbf{AVG} & 95.74 & 73.08 & 72.87 & 61.16 \\
\hline
\end{tabular}

\label{tab:dingo_score}
\end{table}

\subsection{Quality Testing and Validation}
To evaluate the impact of data filtering, we extracted subsets of both the original and filtered datasets for continual pre-training on the LLaMA-3.1 8B model. During the training process, perplexity exhibited a steady downward trend, indicating effective learning. As training progressed, the model's performance on the downstream task MGSM consistently improved. Ultimately, compared to the pre-cleaning dataset, the model trained on the filtered data achieved significantly lower perplexity and superior performance on MGSM. These results demonstrate the enhanced quality of the filtered dataset.

\section{Conclusion}

This study provides a systematic explanation of the construction methodology and data processing workflow of the WanJuanSiLu multilingual open-source corpus. The research team has optimized the data processing workflow, implementing mechanisms such as data cleaning, deduplication, multi-level content safety filtering, and refined quality evaluation. These efforts have culminated in the creation of a large-scale, high-quality, and secure dataset for low-resource languages.

In future research, we will focus on the following areas:
\begin{itemize}
  \item \textbf{Data Processing Architecture Optimization:} Continuously optimizing the data processing pipeline to build a more efficient and scalable architecture.
  \item \textbf{Quality Control System Upgrades:} Iteratively upgrading the quality classification algorithms and content safety models to further enhance the accuracy and reliability of data classification.
  \item \textbf{Language Coverage Expansion:} Expanding the language coverage of WanJuanSiLu by integrating diversified corpus resources to enhance linguistic and cultural diversity.
\end{itemize}

\newpage

\bibliographystyle{unsrt}
\bibliography{wanjuansilu.bib}

\end{document}